\documentclass{article}

\newcommand{\PCignore}[1]{}

\newcommand{\squishlist}{
 \begin{list}{$\bullet$}
  { \setlength{\itemsep}{0pt}
     \setlength{\parsep}{3pt}
     \setlength{\topsep}{3pt}
     \setlength{\partopsep}{0pt}
     \setlength{\leftmargin}{1.5em}
     \setlength{\labelwidth}{1em}
     \setlength{\labelsep}{0.5em} } }

\newcommand{\squishlisttwo}{
 \begin{list}{$\bullet$}
  { \setlength{\itemsep}{0pt}
     \setlength{\parsep}{0pt}
    \setlength{\topsep}{0pt}
    \setlength{\partopsep}{0pt}
    \setlength{\leftmargin}{2em}
    \setlength{\labelwidth}{1.5em}
    \setlength{\labelsep}{0.5em} } }

\newcommand{\squishend}{
  \end{list}  }

\newcommand{\ours}{AQGAN\xspace}

\newcommand{\Q}{Quantizer\xspace}
\newcommand{\Qs}{Quantizers\xspace}

\newcommand{\G}{Generator\xspace}
\newcommand{\D}{Discriminator\xspace}
\newcommand{\direct}{One-step-Q\xspace}
\newcommand{\mobilenet}{MobileNet-V2\xspace}
\newcommand{\resnet}{Resnet-18\xspace}
\newcommand{\resnext}{ResNeXt\xspace}
\newcommand{\mnasnet}{MnasNet\xspace}
\newcommand{\proxyless}{ProxylessNAS\xspace}
\usepackage{microtype}
\usepackage{graphicx}
\usepackage{subfigure}
\usepackage{booktabs} 
\usepackage{amsmath}
\usepackage{mathtools}
\usepackage{graphicx}
\usepackage{comment}
\usepackage[table,xcdraw]{xcolor}
\usepackage{multirow}
\usepackage{microtype}
\usepackage{graphicx}
\usepackage{subfigure}
\usepackage{booktabs} 
\usepackage{amsmath}
\usepackage{mathtools}
\usepackage{graphicx}
\usepackage{comment}
\usepackage{algorithm}
\usepackage{xspace}
\usepackage[table,xcdraw]{xcolor}
\usepackage{multirow}
\usepackage{cite}
\usepackage{amsmath,amssymb,amsfonts}
\usepackage{algorithmic}
\usepackage{textcomp}
\usepackage{xcolor}
\usepackage{amsmath,hyperref}
\usepackage{caption}
\pdfoutput=1

\usepackage{hyperref}

\usepackage[accepted]{sysml2019}

\sysmltitlerunning{Generative Design of Hardware-aware DNNs}

\begin{document}

\twocolumn[
\sysmltitle{Generative Design of Hardware-aware DNNs }



\sysmlsetsymbol{equal}{*}

\begin{sysmlauthorlist}
\sysmlauthor{Sheng-Chun Kao}{to}
\sysmlauthor{Arun Ramamurthy}{goo}
\sysmlauthor{Tushar Krishna}{to}

\end{sysmlauthorlist}

\sysmlaffiliation{to}{Georgia Institute of Technology, GA, USA}
\sysmlaffiliation{goo}{SIEMENS Corporate Technology, NJ, USA}

\sysmlcorrespondingauthor{Sheng-Chun Kao}{felix@gatech.edu}

\sysmlkeywords{Neural Architecture Search, GAN, Quantization, Generative model}

\vskip 0.3in

\begin{abstract}
To efficiently run DNNs on the edge/cloud, many new DNN inference accelerators are being designed and deployed frequently. To enhance the resource efficiency of DNNs, model quantization is a widely-used approach. However, different accelerator/HW has different resources leading to the need for specialized quantization strategy of each HW.
Moreover, using the same quantization for every layer 
may be sub-optimal, increasing the 
design-space of possible quantization choices.
This makes manual- tuning infeasible.
Recent work in automatically determining quantization for 
each layer is driven 
by optimization methods such as reinforcement learning.
However, these approaches need re-training the RL for every new HW platform.

We propose a new way for autonomous quantization and HW-aware tuning. We propose a generative model, \ours, which takes a target accuracy as the condition and generates a suite of quantization configurations. With the conditional generative model, the user can autonomously generate different configurations with different targets in inference time. Moreover, we propose a simplified HW-tuning flow, which uses the generative model to generate proposals and execute simple selection based on the HW resource budget, whose process is fast and interactive. We evaluate our model on five of the widely-used efficient models on the ImageNet dataset. We compare with existing uniform quantization and state-of-the-art autonomous quantization methods. Our generative model shows competitive achieved accuracy, however, with around two degrees less search cost for each design point. Our generative model shows the generated quantization configuration can lead to less than 3.5\% error across all experiments.

\end{abstract}

]



\printAffiliationsAndNotice{}  
\section{Introduction}
\captionsetup[figure]{labelfont=bf}
\captionsetup[table]{labelfont=bf}
Deep Convolution Neural Networks (CNNs) have achieved remarkable accuracy in many problems such as image classification, object detection, and machine translation. These state-of-the-art models are complex and deep, which makes them challenging to implement as a real-time application on edge devices, which are constrained by latency, energy, and model size. Therefore many research directions are being proposed such as hand-crafted efficient DNN models (MobileNet-V2 \cite{sandler2018mobilenetv2}, ResNet50 \cite{he2016deep}) or to quantize the weight and activations of models.

For resource efficiency, researchers have found great success in representing the network with quantized bit-widths~\cite{zhou2016dorefa,courbariaux2014training}. Most conventional
methods 
quantize the model into a uniform bit-width across all layers~\cite{courbariaux2014training}. 
However, since different layers 
exhibit different properties, they have been shown to have different sensitivity to bit-width quantization~\cite{krishnamoorthi2018quantizing}. Thus, researchers have started to quantize layers with different bit-width (layer-wise quantization)~\cite{wang2019haq}.
However, to determine the optimal quantization bit-width for each layer is an extremely complex problem because of the massive design space. The complexity of the design space is composed of number of layers (N), 
each with flexible bit-width (32)\footnote{Assuming the quantization approach supports 1-bit to single precision 32-bit}. For example, a quantized MobileNet-V2 can have a design space of $O(32^{54})$ possible solutions. Therefore many quantization strategies are formulated by rule-based heuristics, and often require domain experts to tune the model. Moreover, this design space is unique for each HW configuration (i.e., a HW with a certain energy, latency, memory budget, and accuracy tolerance), meaning when the underlying HW changes, which happens frequently because of the fast advancement of technology, the entire process has to be repeated again.

A recent work, HAQ \cite{wang2019haq}, utilized a reinforcement learning method to automate the search process, which leaves humans out of the design optimization iteration. To meet different resource constraints, The authors of  HAQ modified the actions made. The HAQ framework permits one to incorporate resource constraints into the reward function as demonstrated by \cite{mnasnet, tan2019efficientnet, kim2017nemo}. However, both methods have some potential challenges: first, the designer needs the expertise to design rules for good action reduction strategy or designing the parameters in the incorporated reward function; second, the HW configuration can alter frequently in practice, and the RL search process needs to relaunch, which leads to large convergence times.

We believe that an ideal scenario when tuning a model for HW in practice is to  have an agent that has learned the mappings from the quantization space to the model accuracy with prior samples such that designers could interact with the educated agent in real-time while tuning the conditions (HW configuration/ resource requirement) flexibly. 

In this paper, we take a key step towards realizing this ideal propose a new autonomous framework for quantizing DNN models called Autonomous Quantization GAN (\ours).
We make two key contributions:
First, we enable 
the generation of quantized networks without the requirement of expert knowledge from both the model and HW perspective;
second, with our proposed \ours, we provide a new simplified HW-aware tuning flow, which leads to the reduction of the search cost for finding the right quantization across different HW resource budgets.


In this work, we define ``response contour" as the  the relationship between accuracy and quantization.
For each DNN model, we learn the
 response contour, and employ several overfitting prevention techniques that we describe later. The response contour of design point (quantization configuration) to accuracy is an n-to-1 mapping, that is, several design model configurations can yield the same accuracy. This opens up the opportunity to leverage the inverse 1-to-n mapping property through a \textit{generative model}. Different HW configurations (e.g., TPU \cite{jouppi2017datacenter}, Eyeriss \cite{chen2016eyeriss}, and ShiDiaNao \cite{du2015shidiannao}) have different HW resource budgets (e.g., memory capacity, bandwidth etc), and hence require different quantization strategies to fit the models. Therefore, we build a conditional GAN (cGAN) based framework, which the designer only needs to specify the conditioned accuracy, and the agent generates a set of different design points for the designer in real-time. With these alternatives, the designer can pick a design point based on the HW configuration at hand. Contrary to RL or optimization-based works, the designer can interact with the agent in real-time with different conditioning accuracy numbers. Following the initial creation of the generative model, no additional training is involved in this interaction. Our proposed  Autonomous Quantization GAN (\ours) enables the generation of quantization configurations by conditioning on the designer's input of accuracy number thus enabling data-driven inverse design process. 

The primary contribution of this paper includes:
\begin{enumerate}
  \item \ours: A new DNN model quantization framework is proposed. Based on the algorithm foundation of cGAN, we build a framework that conditions on a continuous value of model accuracy, to generate a set of quantization configuration. We further enhance the work by training a forward model and incorporating them into the training loop of \ours.
  \item Hardware-aware: We demonstrate how to manage different resource constraints. By referring to the resource consumption model, the designer picks the feasible one and can interact with the agent to explore different solutions.  
  \item Quantitative results: We experiment on various widely-used efficient models, including \mobilenet \cite{sandler2018mobilenetv2}, \resnet \cite{he2016deep}, \resnext \cite{xie2017aggregated}, \proxyless \cite{cai2018proxylessnas}, and \mnasnet \cite{mnasnet}. We compare the performance with conventional uniform quantization algorithm. We compare the performance and search time with the state-of-the-art autonomous quantization method. The experiment shows across all models, resource constraints, and conditioning accuracy, the generated set of quantized models can achieve an accuracy within 3.5\% of that requested.
\end{enumerate}



\section{Related Work}

\subsection{Quantization}
\textbf{Quantization Algorithms.}For resource efficiency, reseachers have found great success in representing the network with quantized bit-width. Deep Compression \cite{han2015deep} quantized the values into bins, where each bin shares the same weight, and only a small number of indices are required. Courbariaux et al., \cite{courbariaux2014training} directly shifted the floating point to fixed point and integer value. 
Many works showed 8-bit can be an empirical quantization strategy, and gysel et al., \cite{gysel2018ristretto} presented a fine-tuning method for quantization. 
To exploit the benefits of low bit-width, DoReFa-Net \cite{zhou2016dorefa} retrained the network after quantization and enabled the quantized backward propagation. 
To find the best quantization at inference time, the quantized-aware training were widely used, in which Krishnamoorthi et al., \cite{krishnamoorthi2018quantizing} quantized the network to hardware-friendly bit-width, and Benoit et al., \cite{int_Quantize} optimized the co-designed training procedure with integer arithmetic inference.
Most  works quantize the model in one-step. \cite{zhou2017incremental} utilizes a multi-step incremental quantization strategy which leads to more accuracy, albeit taking longer.

\textbf{Autonomous flexible quantization.} A fine-grained quantization, layer-wise quantziation could achieve more aggressive quantization. HAQ \cite{wang2019haq} drives the search of the flexible quantization with reinforcement learning DDPG \cite{lillicrap2015continuous} manner. HAQ uses  to figure out the quantization configuration of each layer. HAQ optimizes the configuration to fit the defined target function, which is the Top-1 accuracy difference to the original (full-precision) model. However, the challenges of reinforcement learning method are that they take large number of epochs to converge and they need to be retrained when the target (HW resource budget change). It becomes significant when considering HW-aware quantization in this era of booming new HW platforms.


\subsection{Generative Adversarial Networks (GANs)}
Generative Adversarial Nets (GANs) \cite{goodfellow2014generative} , based on game theory approach, contains a generative network (\textit{generator}) and adversarial network (\textit{discriminator}). The generative model is pitted against the discriminator, which learns to determine whether a sample is from the generator or the training data.
Instead of building multiple GAN networks for different class of images, the fundamental GAN can be augmented by incorporating condition \cite{mirza2014conditional, van2016conditional}. Both the generator and discriminator take in the condition in the training phase, so that the mapping of different condition to different distribution can be learned. Similar approaches such as ACGAN \cite{acgan}, infoGAN \cite{infogan}, and others \cite{odena2016semi} \cite{ramsundar2015massively}, task the discriminator to reconstruct the class label taken by the generator rather than feeding in condition to discriminator. GANs have tackled the tasks of image generation \cite{goodfellow2014generative, infogan, acgan}, image prediction \cite{yoo2016pixel}, text-to-image synthesizing \cite{reed2016generative}, and sequence generation \cite{yu2017seqgan}. Conditional GANs forces the output to be conditioned on the input, and different conditions such as classification label \cite{goodfellow2014generative}, text \cite{reed2016generative}, beauty-level of image \cite{diamant2019beholder}, and image itself \cite{isola2017image} have been applied to different applications. Most of the GAN works has focused on improving and extending the image-based applications such as super-resolution image generation \cite{wang2018esrgan} or image-to-image translation \cite{isola2017image} \cite{almahairi2018augmented}. This work applies GAN to the generation of quantization configuration, conditioned by the user's input of accuracy number.

\section{Approach}
\begin{figure}[]

\begin{center}
\includegraphics[width=0.8\linewidth]{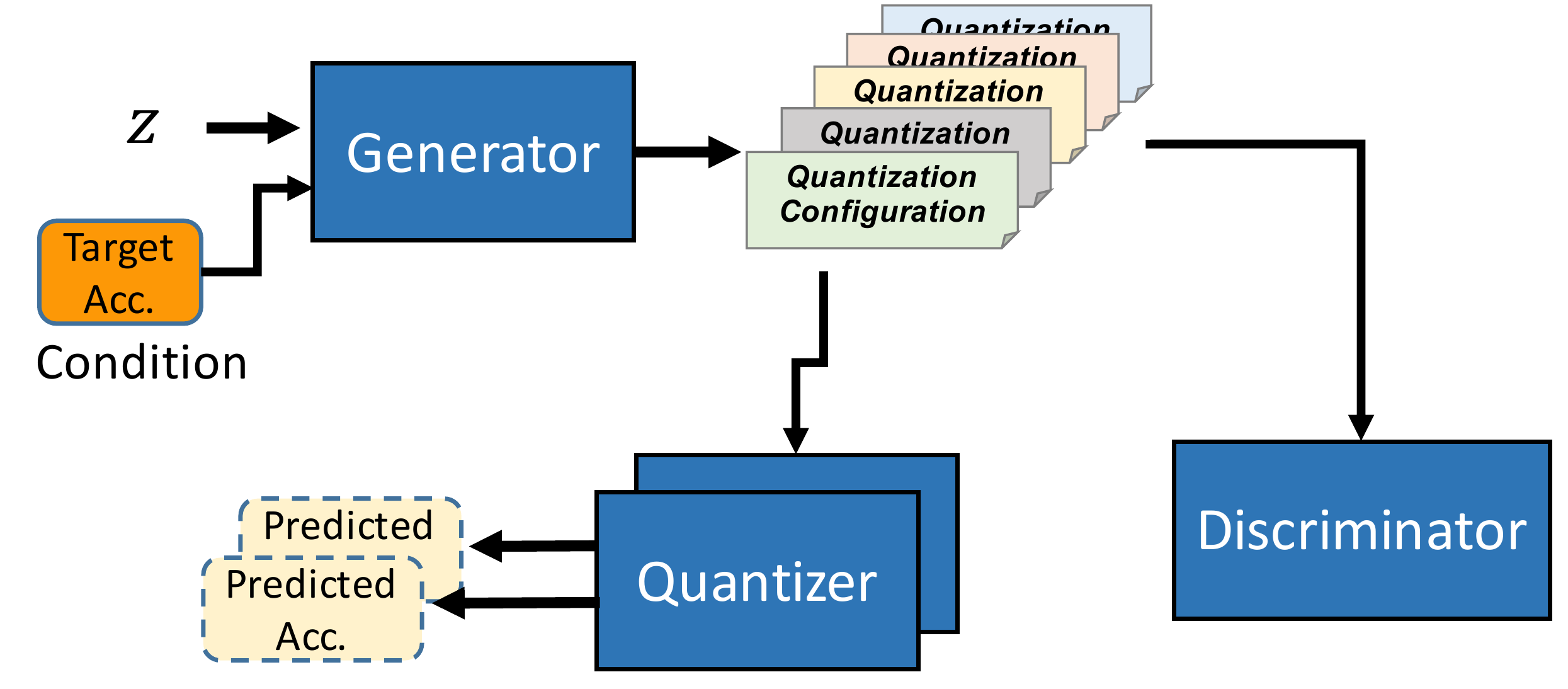}
\end{center}
\vspace{-0.40cm}
  \caption{The stucture of \ours.}

\label{fig:system}
\end{figure}

We model the DNN quantization problem as a generative model training problem. We build a condition-based GAN infrastructure, Autonomous Quantization GAN (\ours), an inverse procedure of the conventional DNN quantization process that conditioning on the targeted accuracy generates a quantization configuration. 
\autoref{fig:system} shows the system overview.
We develop and incorporate several useful techniques to prevent overfitting and mode collapse. We evaluate the generative model with the ground-truth environment and show only 3.5\% error on average. 

\subsection{Experience collection}
The first step of \ours training is to build a ground-truth environment, which is a quantizing procedure of a target model (e.g., quantized MobileNet-V2). On each target model, we apply quantization-aware training to each layer of the target models. We quantize both weights and activations. A complete quantization configuration will describe the bit-width allocation of each layer, which we feed into the built ground-truth model and receive the post-quantized Top-1 accuracy number. This is used to train the generative model in \ours.

An example of a design point (X, Y) is as follows. Assuming a L layer DNN, the X is the quantization configuration, a 1-by-N sequence whose value range from 1 to 32, indicating bit-width. The Y is the corresponding accuracy number of the model being quantized by this X configuration. A design point (X, Y) is one point in the dataset, which we further split into training and testing set. We apply random sampling, which has been proven as a competitive approach to efficiently sample the through the response contour \cite{such2017deep}, and generate a dataset of design point to response pair. With the dataset at hand, we apply the 80-20 split to formulate the training and testing set. The testing set is never seen in any training process of \G, \D, or \Q, which we will describe later. In the vast design space (more than billions of design points), we found that the designed model can achieve competitive testing performance by the left-out testing set while we only collect 20K design points\footnote{We empirically find 20K design points are enough (where the marginal performance gain start to flatten) for the models we investigated.} as a dataset in each ground-truth environment of modern models, including \mobilenet, \resnet, \resnext, \proxyless, and \mnasnet.

\subsection{Condition labeling}
With the training set at hand, we label the data by normalizing the response accuracy number to the continuous range of [0, 1]. The labels are taken by \ours as a condition for generating the quantization configuration.

The designer with a well-trained generator on a target model will serve the targeted accuracy number to \ours as an input. \ours normalizes the targeted accuracy to its label range [0, 1], condition on this label, and generates the corresponding quantization configuration.

\subsection{\ours structure}
We train our GAN infrastructure with two-step training. First, we train the \Q, an accuracy-predicting model that serves as an auxiliary instructor in the next step, which is inspired by the auxiliary classifier in ACGAN \cite{acgan} and the classification of a discriminator in infoGAN\cite{infogan}. Second, we train the generator and discriminator with \Q as an auxiliary instructor.

\subsubsection{\Q.}
The \Q is built as an MLP model. We train \Q with the training dataset as a regression problem, which takes the design points (quantization configuration) and regresses the value of the corresponding label, i.e., the normalized accuracy number. In the evaluation phase, \Q takes in a design point in the testing set and predicts its corresponding label. We evaluate the performance of \Q by the L1-loss to the true label, and our model achieves under 3.5\% error across all the DNN models considered. The intuition of our design is that we first stabilized the predictor before the iterative discriminator and generator training loop to relieve the stress of discriminator, which usually plays a crucial role in the convergence of GAN.

\subsubsection{Overfitting prevention.}
When training the \Q, we include conventional overfitting prevention methods such as batch normalization, dropout, and early stopping to strike the balance between variance and bias owing to the difference of model complexity. In addition, we use an ensemble method comprising of multiple \Qs to guide the training of the generative model. We train multiple \Qs with different complexity from thin MLP to wide MLP. In the experiment, we train 4 \Qs, each with three layers and with 64, 128, 256, 512 nodes on each layer respectively. The number of \Qs is decided by empirical experiment, where we found 4 \Qs was sufficient. They are trained individually with the same training set. The intuition is that the thin MLPs can capture more general feature while the wide MLPs are capturing a sophisticated one. After training, the parameters of this group of \Qs are fixed and they will be the group of instructors for the generator in the next step.

\subsubsection{\G.}
The training of \G and \D starts at the second step. The \G is built with MLP infrastructure with $L$ output dimension and $D+C$ input dimension, where $L$ is the length of quantization configuration that needs to estimated which is also the number of layers/blocks of the model that we are quantizing; $D$ is the dimension of the latent space, which is a $D$ dimension gaussian noise, $N(0,1)$, and we take $D=10$ in our experiments; $C$ is the dimension of condition, where we have $C=1$ because the design of one continuous label as condition. When training \G,  we sample a batch of fake labels in the range of [0, 1] as the condition and a batch of noise in $N(0,1)$ as latent space and which is fed to \G. We the collect the $L$ dimension outputs, which is a fake quantization configuration.  Assuming quantizing a DNN with $L$ layer, the output space of the generative model is a 1-by-N sequence, whose values are range from 1 to 32, indicating the bits-width.

\subsubsection{\D.}
\D is also built with MLP infrastructure with $1$ output dimension and $L$ input dimension, where $1$ is the judgment of true or fake; $L$ is the same dimension as \G outputs. \D is trained as a standard discriminator, which takes the $L$ dimension design points from the training set and the $L$ dimension outputs from \G and are tasked to judge whether they are true (from the training set) or fake (from \G).

\subsection{GAN training}
In the second step of training, the parameters of \Qs are not updated. We use three kinds of loss in our framework. \D , \G  and \Qs losses to estimate the parameters of the \D and the \G.
\subsubsection{\Qs loss:}
The kernel part of the framework that drives the learning process is the next \Qs loss. In each iteration of training, we use a batch of fake label $\bar{Y}$ to let \G output fake data. We feed the generated fake data into the group of \Qs, which predict the corresponding label value $\bar{Y_{Q1}}$, $\bar{Y_{Q2}}$...,$\bar{Y_{QN}}$, where $N=5$ in the our setting. \Qs together play the instructors' role to teach \G to generate fake data that can lead to the good predicted label value. We use mean square error (MSE) loss. The \Q loss $L_{Q}$ is the sum of each loss term as follow: 
\begin{align}
\hskip\parindent & \begin{gathered}
    L_{Q} = \sum l_{Qn} = \sum l(\bar{Y}, \bar{Y_{Qn}})= \sum MSE(\bar{Y}, Q_{n}(G(\bar{Y}))))
    \end{gathered} &
\end{align}
It is worth noting that we have $N$ corresponding predicted labels $\bar{Y_{Qn}}$ with the same $Y$ and same \G. We let each of these instruct the \G to prevent \G overfitting to any single \Q. Also, the wider \Q instruct \G to deal with sophisticated generation task, and the thinner \Q regularized \G to build more general generation rule.

\textbf{\D and \G loss:} \D loss is the performance of classifying the training set from the fake set to true and fake, and \G loss is the performance of \G making \D misclassify. \D and \G loss are defined as in a conventional GAN. We compute the loss with Wasserstein distance and leverage gradient penalty to stabilize the training and prevent mode collapse in our experiments.

With the three training losses defined, we train the model iteratively in this second training step. The insight of this framework is that with a good pre-trained instructor, the stress on the \D is relieved so that the \D can keep pace with the improvements of \G, thus avoiding to mode collapse.

\subsection{Generative process and evaluation}
After \ours is trained, the designer can use the \G as a conditional generative model.
\subsubsection{Evaluation and performance index.} The model is trained on the ground-truth environment, where the target model is defined and the parameters initialization seed is set. During testing, we randomly sample a batch of target accuracy $\bar{y}$ and normalize them into label domain to be between [0, 1], noted as $\bar{Y}$ in the context. We feed those label into \ours as condition and collect the batch of output from \G, which are fake configuration (generated quantization configuration). We evaluate those fake configuration by testing them in the ground-truth environment, i.e., we apply those configurations to quantize the target DNN  model and gather the corresponding Top-1 accuracy, which is the ground-truth accuracy ($y$). We use L1-loss to measure the performance index of the model. The loss of generative model, $L_{model}$, is defined as,
\begin{align}
\hskip\parindent & \begin{gathered}
    L_{model} = \frac{1}{K} \sum_{i=1}^{K}(\left | y_{i} - \bar{y} \right |)
    \end{gathered} &
\label{formula:L1}
\end{align}
, where K is the batch size.
\subsubsection{Generative design process.}
 The designer specifies a target accuracy as condition to the generative model and the expected batch size to be generated as outputs. The accuracy is normalized internally and \ours produces a batch of outputs, which are the set of valid quantization configuration predicted to meet the specified condition.

\subsubsection{Implementation details}
The backbone of \Q, \G, and \D are MLP with dropout and batch normalization. We apply drop out rate of 50\% and utilize a batch size of 256. For \G, we use the latent space of 10. We use Westeran distance as our loss function.


\subsection{Autonomous quantization}
The benefits of having a generative model making proposals of designs are two fold: First, is the automation of finding flexible quantization bit-width of each layer. Second, is the simplified HW-aware tuning process.

\subsubsection{Autonomous flexible quantization}
Owing to the sensitivity of each layer, an uniform quantization implementation of the DNN may not be suffient to meet the accuracy constraints sought by the designer. To flexibly search the design space, an optimal quantization configuration is needed for each layer of the DNN. However, as the search space is massive it is not possible to exhaustively search for the optimal answer. Hence, some intelligent autonomous methods such RL, HAQ have been designed. 

Our work, a generative model, is also a framework of generating the quantization configuration autonomously. Taking in the target accuracy, our work generates a set of design proposals that meets the accuracy targetted. The difference between HAQ (RL-based method) and ours (generative model) is the search cost and the number of designs that can be generated with each search. The RL-based method could take hundreds to thousands of epochs to converge to a single design point. However, with a generative model, the multiple design proposals are generated instantaneously during inference. While the RL searches and train simultaneously the generative model relies on previously gathered data, but once trained the model can generate any number of design points during inference and for different accuracy conditions.

\subsubsection{HW-aware tuning}
The goal of quantization is to create an HW-efficient DNN model, having good property of HW performance such as memory, energy, or latency. Thus, quantization of a DNN model occurs with a specification of the HW budget (resource). In an RL-based method, HW performance index can be incorporated into the target function to drive the search process or the actions recommended can be modified as in HAQ. However, both these approaches require a pre-defined HW budget. The search is, thus, driven by the target function (accuracy number) and the HW budget. However, it is often the case that the HW budget could change dynamically, such as the energy constraints of different systems or different memory capabilities in different platforms. Traditional approaches require a new search for each different HW consideration limiting the ability to transfer any knowledge gained.

To overcome this limitation, we propose a new simplified design flow for HW-aware tuning. A trained generative model, generates a set of proposals with similar accuracy number but different HW requirements. Our HW-aware tuning process is as follows: First, the model generates a set of design proposals based on the designer's desired target accuracy. Second, the designer ranks the proposals with the HW performance that is to be optimized. Finally, the designer selects one of them that fits the HW budget the best. These three steps could be accomplished with a simple and fast program, which is only executes sorting of the generated design alternatives and selects one or more that meets the hardware constraints.

With our \ours -based HW-aware tuning flow, the generative model could be distributed to different designers irrespective of their application, i.e., both a ML practitioner working on IoT devices and a ML practitioner working on cloud acceleration would utilize the same generative model. The only difference, is they run the HW-aware tuning flow with different HW budget / constraints.

\section{Experiments}
\begin{table*}[t]
\centering
\caption{Comparisons with One-step-Q Quantization}
\includegraphics[width=1\linewidth]{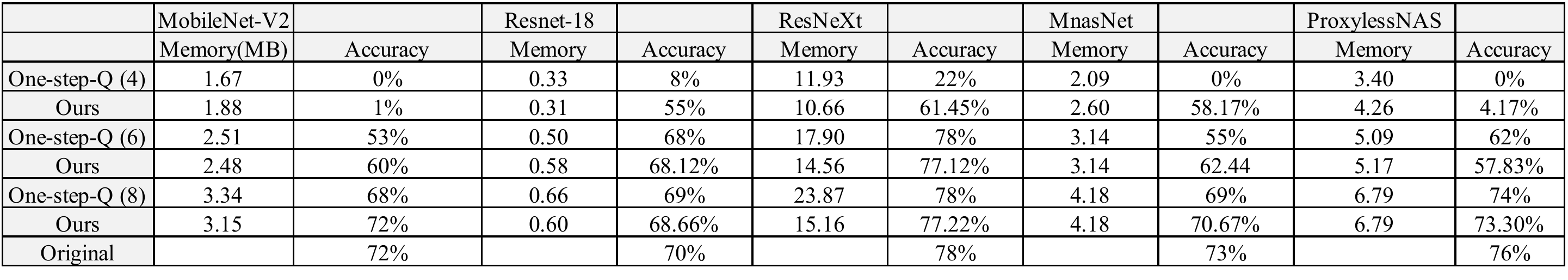}
\vspace{-0.5cm}
\label{table1}
\end{table*}

\subsection{Datasets and models}
As our focus is to provide flexible bit-width allocation for different layers, we apply standard range-based linear quantization\footnote{We pick the scaling factor based on the range of the current tensor (weight/ activations). We pick the scaling factor by $S = 2^{bit}/(max_{tensor} - min_{tensor})$, where $bit$ is the quantization bit-width. The $max_{tensor}$ and $min_{tensor}$ are the maximum and minimum values in the tensor. The quantization formula becomes $v_{int} = (v_{float} - min_{tensor}) \times S - 2^{bit-1}$. The $v_{int}$ and $v_{float}$ indicate a post-quantized and pre-quantized value in a tensor.} for each layer. There are advanced quantization algorithms such as DoReFa \cite{zhou2016dorefa} with low bandwidth gradient, PACT \cite{choi2018pact} with parameterized clipping activation, WRPN \cite{mishra2017wrpn} with wide reduced-precision, QIL \cite{qil}, and HAWQ-V2 \cite{hawq}. These papers work on quantization algorithm, while the bit-width need to be exhaustively searched, manual-assigned, or assigned by heuristic. In this work, we have an orthogonal goal. Our focus is to automate the process of bit-width assignment, and we apply the basic ranged-based quantization scheme. However, two techniques can be combined for the extension of this work to reach a better result. HAQ \cite{wang2019haq}, which uses RL to automate the process and uses the basic quantization scheme, is the most closely related work to ours. Therefore we pick HAQ as our main comparison.

\subsection{Comparisons of related works}
\subsubsection{One-step quantization.}
First, we compare our works with the one-step quantization methods as applied in \cite{chen2015compressing, gupta2015deep} with uniform bit-width, and we name this as \direct. The \direct is performed with the same procedure of our framework, except that they are quantized with uniform bit-width.
\subsubsection{Autonomous quantization}
The performances of our autonomous quantization method is compared against HAQ, which also searches for the quantization bit-width of each layer. We apply the same setting as in the original paper with the modification that HAQ is able to target different accuracy.

\subsection{Memory-constrained quantization}
We construct the experiments such that the hardware considered are constrained by the DNN parameter memory computed post-quantization. We compare our results to \direct, which quantizes the model instantly without need for expensive searches. We apply 8-bit, 6-bit, and 4-bit quantization by \direct to quantize five different models and list their parameter memory size and the model accuracy in \autoref{table1}. Then we examine under similar parameter memory constraints, the accuracy \ours can achieve. The flows is as follows:

The generator proposes the set of model configurations when constrained by the desired target accuracy. Following this, we could apply the HW-aware tuning flow of \ours, i.e., we pick the designs that are closest to the memory size of the \direct method and report the accuracy that \ours targets. We also evaluate the maximum accuracy that can be achieved given a resource constraint. This is realized by searching through the possible set of accuracies for designs that meet a specified resource constraint.

From \autoref{table1}, it is seen that given a memory constraint, our work leads to of designs with higher accuracy. In very few cases, our approach degrades the accuracy by less than 1\%. The effectiveness of our works is more apparent when it comes to low bit-width. In 4-bit cases, \direct has bad accuracy, where sometimes the model has an accuracy of 0\%. However, with flexible bit-assignment, in most cases, our work can generate a quantized model with much higher accuracy. It is worth noted that unlike DoReFa \cite{zhou2016dorefa}, PACT \cite{choi2018pact}, WRPN \cite{mishra2017wrpn}, QIL \cite{qil}, HAWQ-V2 \cite{hawq}, whose bit-width strategy need to be exhaustively searched or manually-assigned, and HAQ \cite{wang2019haq}, whose bit-width strategy need a search process through RL algorithm, \ours generates the bit-width strategy with generative model at inference time without retraining.

\begin{table*}[t]
\centering
\caption{Comparisons with HAQ: MnasNet}
\includegraphics[width=1\linewidth]{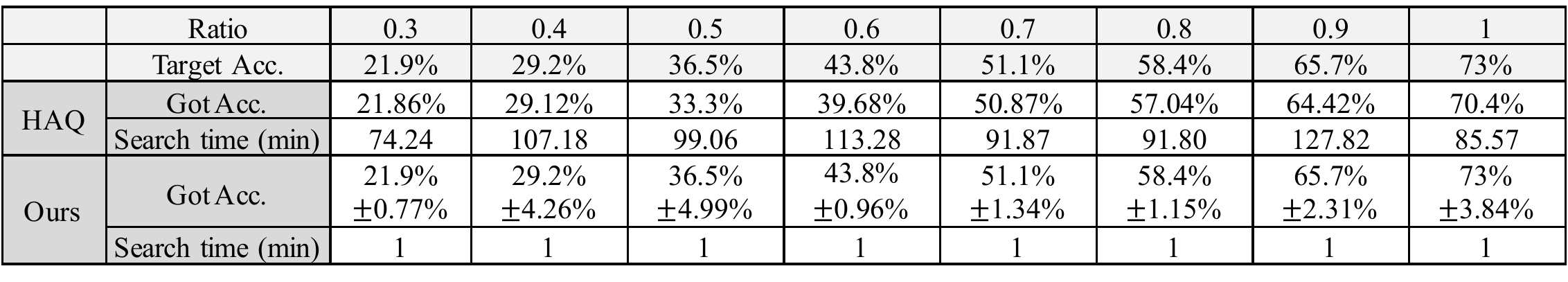}
\vspace{-0.5cm}
\label{table_haq_mnasnet}
\end{table*}

\begin{table*}[t]
\centering
\caption{Comparisons with HAQ: MobilenetV2}
\includegraphics[width=1\linewidth]{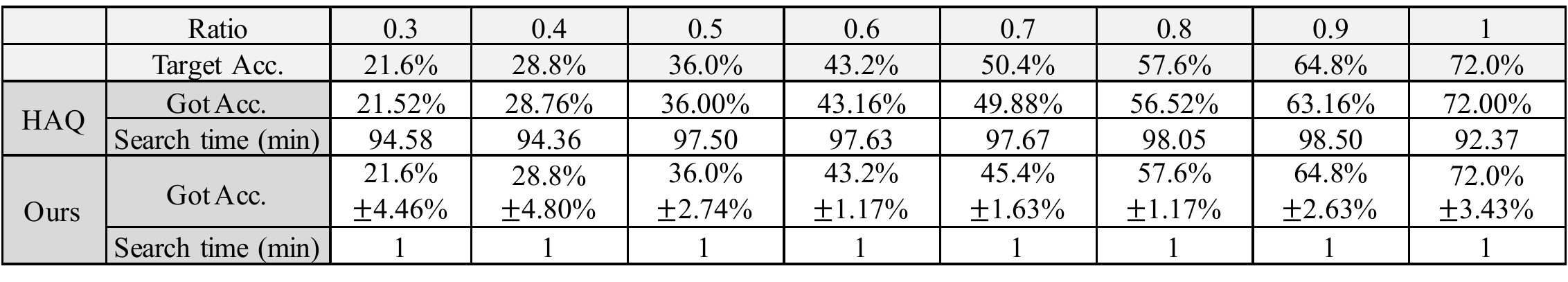}
\vspace{-0.5cm}
\label{table_haq_mobilenet}
\end{table*}

\begin{table*}[h]
\centering
\caption{Overall Performance of AQGAN}
\includegraphics[width=1\linewidth]{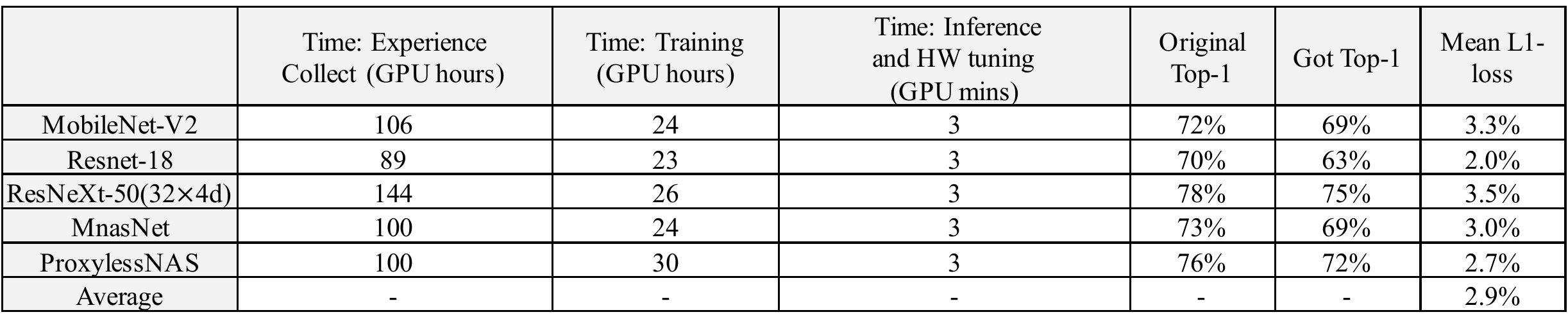}
\vspace{-0.5cm}
\label{our_table}
\end{table*}

\begin{figure}[h]
\begin{center}
\includegraphics[width=1\linewidth]{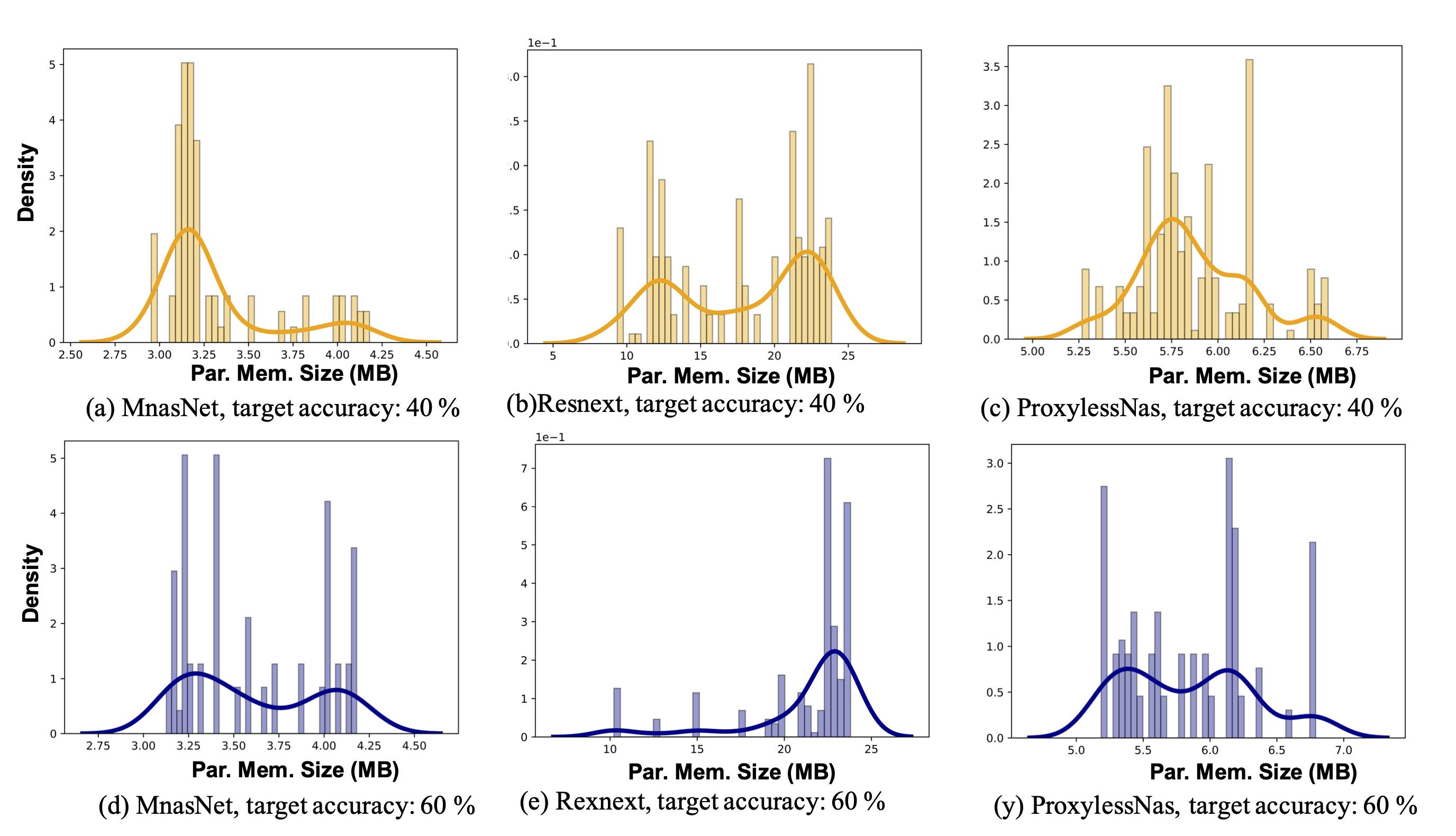}
\end{center}
\vspace{-0.40cm}
  \caption{The distribution of the parameter memory size in different model, conditioning on accuracy number:40\% (top, yellow) and 60\% (bottom, blue).}

\label{fig:hist_p}
\end{figure}

\begin{figure}[h]

\begin{center}
\includegraphics[width=1\linewidth]{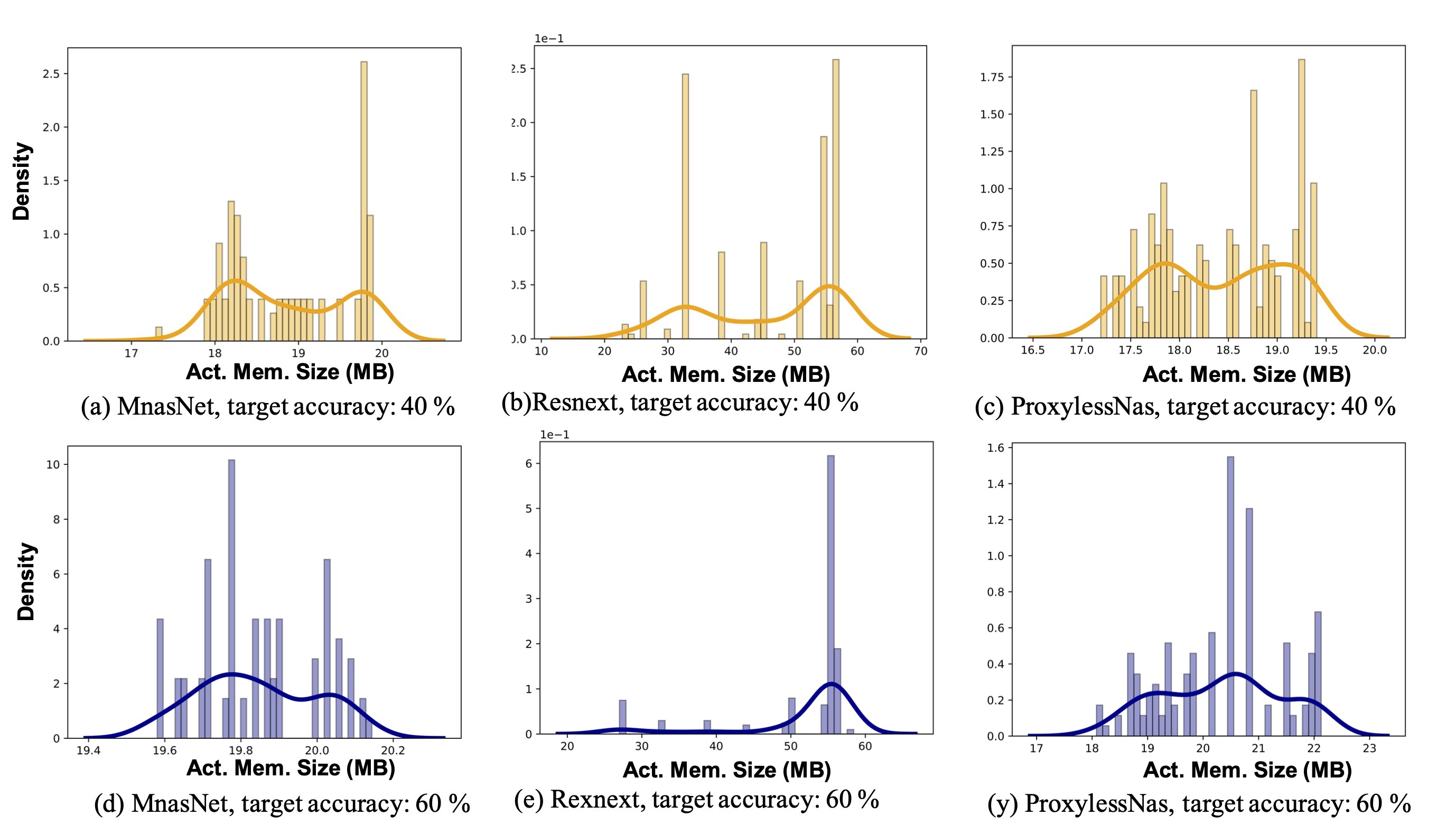}
\end{center}
\vspace{-0.40cm}
  \caption{The distribution of the activation memory size in different model, conditioning on accuracy number:40\% (top, yellow) and 60\% (bottom, blue).}

\label{fig:hist_a}
\end{figure}

\subsection{Autonomous flexible quantization}
We construct experiments that perform autonomous flexible quantization and target different accuracies. We compare it with HAQ. We construct the target function to target different accuracy numbers being 0.3, 0.4,..., 1 times that of the original accuracy. 

\textbf{Experiment detail: HAQ.} As indicated by \autoref{table_haq_mnasnet}, there are 8 tasks (columns) for each algorithm (HAQ, \ours). For each task, the algorithm is required to quantize the model such that it is close to the target accuracy number (e.g., target accuracy=$21.9\%$ for the first task in \autoref{table_haq_mnasnet}). For HAQ, we train it by modifying the reward function. Rather than always maximizing the accuracy (which corresponds to one of the tasks of our experiments – the result in the last task (column)), we define the reward to minimize the distance between target accuracy and achieved accuracy. 

\textbf{Experiment detail: \ours.} For our work (conditional GAN-based), we simply use the target accuracy as the condition, sample 50 points from normal distribution $N(0,1)$ as noise, and generate 50 outputs (quantization configuration). We then evaluate the accuracy of each output and gather the accuracy number of the 50 quantization configurations. We calculate their distances to the target accuracy as error range. Therefore, the recorded values become the target value $\pm$ the error range. For example, for the first column in \autoref{table_haq_mnasnet}, the error range is $\pm 0.77\%$.

\textbf{Discussion.} According to \autoref{table_haq_mnasnet}, HAQ can achieve a model with its accuracy number close to target accuracy in most of the cases. Our work generates a set of model configurations that are within close margins to the accuracy requested as recorded in the table. Both works perform well when targeting the accuracy number. However, for each experiment of target accuracy, HAQ needs about 98 GPU mins to complete the search task, while our work needs less than 1 GPU mins to execute an inference as a generative model. We also conduct a similar experiment on Mobilenet-V2. The search cost for HAQ is about 96 GPU mins, which our work stays under 1 GPU mins.

The search time cost play a significant role when ML practitioners deploys models on different platforms that targeting different use cases with different resource budgets such as energy/latency, where the required accuracy level is different. The search cost is the time they need to pay for a new design point in a new hardware configuration.

\subsection{The effectiveness of \ours}
In this section, we inspect the effectiveness of the set of design proposals generated by our model. In the evaluation, we condition on different target accuracy number and gather the quantization configuration that our model generates. We collect the statistic of the parameter memory size of those proposals. The distribution of their model sizes is shown in \autoref{fig:hist_p}. The same condition accuracy leads to a bag of a proposal that has a range of different parameter memory sizes. Those are the generated design points (inference result) the ML practitioner can choose from. Similarly, we show the distribution of the associated activation memory size for different target accuracies in \autoref{fig:hist_a}. Likewise, we could also examine their distribution of latency, energy, or other HW performance that the ML practitioners are targeting. With this generative model at hand, the only task ML practitioners are left is to rank them quickly and select one or more of them.

\subsection{In-depth evaluation of \ours}
We evaluate how our works performs on average in all the five models we apply to. We show the time for each phase of training. We examine two performance index the first being the Top-1 accuracy that the generated model configurations can achieve and the second, is the L1-loss described in \autoref{formula:L1}. We measure the L1-loss, \label{formula:L1}, by random sampling a batch of desired accuracies. We use the generative model to generate a set of quantization configurations which are applied to the ground-truth model, thus generating the evaluation data. Finally, we compute the L1-loss by comparing the ground-truth accuracy with that requested by the designer, i.e., the target accuracy. 

From \autoref{our_table}, we show the time taken for each phase of the generative model. The experience collection takes about 108 GPU hours, and training takes about 25 GPU hours. However, these are both one-time cost, once the experience is collected and the generative models are trained, we could use the same model in different hardware settings to automatically generate quantization configuration or executing HW-aware tuning.

For the top-1 accuracy, we could observe there is a samll accuracy drop. It is because of the fact that we apply one-step training and with basic range-based linear quantization algorithm. A better Top-1 accuracy could be achieved when we apply more advanced methods on those two aspects. However, this work is to discuss the effectiveness of the generative model approach and its fast HW-aware tuning flow. Hence, we leave the evaluation of advanced techniques as our future work.

For the mean L1-loss, we observe that each model reports a loss of under 3.5\% and on average 2.9\% across five models. 

\textbf{Explanation of the results:} When the designer is given a trained \ours and they condition on a target accuracy number, the model generates a set of quantization configurations whose mean L1-loss to the target accuracy number is under 3\%. The designer can follow the HW-aware tuning flow to pick one of them based on the resource constraints at hand.

\section{Conclusion}
Quantization is a key technique for enhancing energy-efficiency of DNN models inside accelerators. Unfortunately, the design-space of 
optimal quantization values for each layer
is extremely large as it depends on the number of layers in the model and the number of levels of quantization the hardware can support.
Moreover, 
 multiple quantization configurations with the same accuracy can have vastly different memory requirements, making it challenging to design an automated framework to find the 
 right quantization values for the specific hardware platform.

In this work, we propose a generative design method, \ours, to autonomously quantize  DNN models. 
Our key novelty is to leverage conditional GANs to learn the 
relationship between 
accuracy and quantization, and 
generate quantization values for a given target accuracy.
Leveraging \ours, we propose a new simplified HW-aware tuning flow to enable rapid HW-aware DNN deployment - both on ASICs and FPGA platforms.

\nocite{langley00}

\bibliography{main}

\begin{thebibliography}{40}
\providecommand{\natexlab}[1]{#1}
\providecommand{\url}[1]{\texttt{#1}}
\expandafter\ifx\csname urlstyle\endcsname\relax
  \providecommand{\doi}[1]{doi: #1}\else
  \providecommand{\doi}{doi: \begingroup \urlstyle{rm}\Url}\fi

\bibitem[Almahairi et~al.(2018)Almahairi, Rajeswar, Sordoni, Bachman, and
  Courville]{almahairi2018augmented}
Almahairi, A., Rajeswar, S., Sordoni, A., Bachman, P., and Courville, A.
\newblock Augmented cyclegan: Learning many-to-many mappings from unpaired
  data.
\newblock \emph{arXiv preprint arXiv:1802.10151}, 2018.

\bibitem[Cai et~al.(2018)Cai, Zhu, and Han]{cai2018proxylessnas}
Cai, H., Zhu, L., and Han, S.
\newblock Proxylessnas: Direct neural architecture search on target task and
  hardware.
\newblock \emph{arXiv preprint arXiv:1812.00332}, 2018.

\bibitem[Chen et~al.(2015)Chen, Wilson, Tyree, Weinberger, and
  Chen]{chen2015compressing}
Chen, W., Wilson, J., Tyree, S., Weinberger, K., and Chen, Y.
\newblock Compressing neural networks with the hashing trick.
\newblock In \emph{International conference on machine learning}, pp.\
  2285--2294, 2015.

\bibitem[Chen et~al.(2016{\natexlab{a}})Chen, Duan, Houthooft, Schulman,
  Sutskever, and Abbeel]{infogan}
Chen, X., Duan, Y., Houthooft, R., Schulman, J., Sutskever, I., and Abbeel, P.
\newblock Infogan: Interpretable representation learning by information
  maximizing generative adversarial nets.
\newblock In \emph{Advances in neural information processing systems}, pp.\
  2172--2180, 2016{\natexlab{a}}.

\bibitem[Chen et~al.(2016{\natexlab{b}})Chen, Krishna, Emer, and
  Sze]{chen2016eyeriss}
Chen, Y.-H., Krishna, T., Emer, J.~S., and Sze, V.
\newblock Eyeriss: An energy-efficient reconfigurable accelerator for deep
  convolutional neural networks.
\newblock \emph{IEEE journal of solid-state circuits}, 52\penalty0
  (1):\penalty0 127--138, 2016{\natexlab{b}}.

\bibitem[Choi et~al.(2018)Choi, Wang, Venkataramani, Chuang, Srinivasan, and
  Gopalakrishnan]{choi2018pact}
Choi, J., Wang, Z., Venkataramani, S., Chuang, P. I.-J., Srinivasan, V., and
  Gopalakrishnan, K.
\newblock Pact: Parameterized clipping activation for quantized neural
  networks.
\newblock \emph{arXiv preprint arXiv:1805.06085}, 2018.

\bibitem[Courbariaux et~al.(2014)Courbariaux, Bengio, and
  David]{courbariaux2014training}
Courbariaux, M., Bengio, Y., and David, J.-P.
\newblock Training deep neural networks with low precision multiplications.
\newblock \emph{arXiv preprint arXiv:1412.7024}, 2014.

\bibitem[Diamant et~al.(2019)Diamant, Zadok, Baskin, Schwartz, and
  Bronstein]{diamant2019beholder}
Diamant, N., Zadok, D., Baskin, C., Schwartz, E., and Bronstein, A.~M.
\newblock Beholder-gan: Generation and beautification of facial images with
  conditioning on their beauty level.
\newblock \emph{arXiv preprint arXiv:1902.02593}, 2019.

\bibitem[Dong et~al.(2019)Dong, Yao, Cai, Arfeen, Gholami, Mahoney, and
  Keutzer]{hawq}
Dong, Z., Yao, Z., Cai, Y., Arfeen, D., Gholami, A., Mahoney, M.~W., and
  Keutzer, K.
\newblock Hawq-v2: Hessian aware trace-weighted quantization of neural
  networks.
\newblock \emph{arXiv preprint arXiv:1911.03852}, 2019.

\bibitem[Du et~al.(2015)Du, Fasthuber, Chen, Ienne, Li, Luo, Feng, Chen, and
  Temam]{du2015shidiannao}
Du, Z., Fasthuber, R., Chen, T., Ienne, P., Li, L., Luo, T., Feng, X., Chen,
  Y., and Temam, O.
\newblock Shidiannao: Shifting vision processing closer to the sensor.
\newblock In \emph{Proceedings of the 42nd Annual International Symposium on
  Computer Architecture}, pp.\  92--104, 2015.

\bibitem[Goodfellow et~al.(2014)Goodfellow, Pouget-Abadie, Mirza, Xu,
  Warde-Farley, Ozair, Courville, and Bengio]{goodfellow2014generative}
Goodfellow, I., Pouget-Abadie, J., Mirza, M., Xu, B., Warde-Farley, D., Ozair,
  S., Courville, A., and Bengio, Y.
\newblock Generative adversarial nets.
\newblock In \emph{Advances in neural information processing systems}, pp.\
  2672--2680, 2014.

\bibitem[Gupta et~al.(2015)Gupta, Agrawal, Gopalakrishnan, and
  Narayanan]{gupta2015deep}
Gupta, S., Agrawal, A., Gopalakrishnan, K., and Narayanan, P.
\newblock Deep learning with limited numerical precision.
\newblock In \emph{International Conference on Machine Learning}, pp.\
  1737--1746, 2015.

\bibitem[Gysel et~al.(2018)Gysel, Pimentel, Motamedi, and
  Ghiasi]{gysel2018ristretto}
Gysel, P., Pimentel, J., Motamedi, M., and Ghiasi, S.
\newblock Ristretto: A framework for empirical study of resource-efficient
  inference in convolutional neural networks.
\newblock \emph{IEEE Transactions on Neural Networks and Learning Systems},
  29\penalty0 (11):\penalty0 5784--5789, 2018.

\bibitem[Han et~al.(2015)Han, Mao, and Dally]{han2015deep}
Han, S., Mao, H., and Dally, W.~J.
\newblock Deep compression: Compressing deep neural networks with pruning,
  trained quantization and huffman coding.
\newblock \emph{arXiv preprint arXiv:1510.00149}, 2015.

\bibitem[He et~al.(2016)He, Zhang, Ren, and Sun]{he2016deep}
He, K., Zhang, X., Ren, S., and Sun, J.
\newblock Deep residual learning for image recognition.
\newblock In \emph{Proceedings of the IEEE conference on computer vision and
  pattern recognition}, pp.\  770--778, 2016.

\bibitem[Isola et~al.(2017)Isola, Zhu, Zhou, and Efros]{isola2017image}
Isola, P., Zhu, J.-Y., Zhou, T., and Efros, A.~A.
\newblock Image-to-image translation with conditional adversarial networks.
\newblock In \emph{Proceedings of the IEEE conference on computer vision and
  pattern recognition}, pp.\  1125--1134, 2017.

\bibitem[{Jacob} et~al.(2017){Jacob}, {Kligys}, {Chen}, {Zhu}, {Tang},
  {Howard}, {Adam}, and {Kalenichenko}]{int_Quantize}
{Jacob}, B., {Kligys}, S., {Chen}, B., {Zhu}, M., {Tang}, M., {Howard}, A.,
  {Adam}, H., and {Kalenichenko}, D.
\newblock {Quantization and Training of Neural Networks for Efficient
  Integer-Arithmetic-Only Inference}.
\newblock \emph{arXiv e-prints}, art. arXiv:1712.05877, Dec 2017.

\bibitem[Jouppi et~al.(2017)Jouppi, Young, Patil, Patterson, Agrawal, Bajwa,
  Bates, Bhatia, Boden, Borchers, et~al.]{jouppi2017datacenter}
Jouppi, N.~P., Young, C., Patil, N., Patterson, D., Agrawal, G., Bajwa, R.,
  Bates, S., Bhatia, S., Boden, N., Borchers, A., et~al.
\newblock In-datacenter performance analysis of a tensor processing unit.
\newblock In \emph{Proceedings of the 44th Annual International Symposium on
  Computer Architecture}, pp.\  1--12, 2017.

\bibitem[Jung et~al.(2019)Jung, Son, Lee, Son, Han, Kwak, Hwang, and Choi]{qil}
Jung, S., Son, C., Lee, S., Son, J., Han, J.-J., Kwak, Y., Hwang, S.~J., and
  Choi, C.
\newblock Learning to quantize deep networks by optimizing quantization
  intervals with task loss.
\newblock In \emph{Proceedings of the IEEE Conference on Computer Vision and
  Pattern Recognition}, pp.\  4350--4359, 2019.

\bibitem[Kim et~al.(2017)Kim, Reddy, Yun, and Seo]{kim2017nemo}
Kim, Y.-H., Reddy, B., Yun, S., and Seo, C.
\newblock Nemo: Neuro-evolution with multiobjective optimization of deep neural
  network for speed and accuracy.
\newblock In \emph{ICML 2017 AutoML Workshop}, 2017.

\bibitem[Krishnamoorthi(2018)]{krishnamoorthi2018quantizing}
Krishnamoorthi, R.
\newblock Quantizing deep convolutional networks for efficient inference: A
  whitepaper.
\newblock \emph{arXiv preprint arXiv:1806.08342}, 2018.

\bibitem[Lillicrap et~al.(2015)Lillicrap, Hunt, Pritzel, Heess, Erez, Tassa,
  Silver, and Wierstra]{lillicrap2015continuous}
Lillicrap, T.~P., Hunt, J.~J., Pritzel, A., Heess, N., Erez, T., Tassa, Y.,
  Silver, D., and Wierstra, D.
\newblock Continuous control with deep reinforcement learning.
\newblock \emph{arXiv preprint arXiv:1509.02971}, 2015.

\bibitem[Mirza \& Osindero(2014)Mirza and Osindero]{mirza2014conditional}
Mirza, M. and Osindero, S.
\newblock Conditional generative adversarial nets.
\newblock \emph{arXiv preprint arXiv:1411.1784}, 2014.

\bibitem[Mishra et~al.(2017)Mishra, Nurvitadhi, Cook, and Marr]{mishra2017wrpn}
Mishra, A., Nurvitadhi, E., Cook, J.~J., and Marr, D.
\newblock Wrpn: wide reduced-precision networks.
\newblock \emph{arXiv preprint arXiv:1709.01134}, 2017.

\bibitem[Odena(2016)]{odena2016semi}
Odena, A.
\newblock Semi-supervised learning with generative adversarial networks.
\newblock \emph{arXiv preprint arXiv:1606.01583}, 2016.

\bibitem[Odena et~al.(2017)Odena, Olah, and Shlens]{acgan}
Odena, A., Olah, C., and Shlens, J.
\newblock Conditional image synthesis with auxiliary classifier gans.
\newblock In \emph{Proceedings of the 34th International Conference on Machine
  Learning-Volume 70}, pp.\  2642--2651. JMLR. org, 2017.

\bibitem[Ramsundar et~al.(2015)Ramsundar, Kearnes, Riley, Webster, Konerding,
  and Pande]{ramsundar2015massively}
Ramsundar, B., Kearnes, S., Riley, P., Webster, D., Konerding, D., and Pande,
  V.
\newblock Massively multitask networks for drug discovery.
\newblock \emph{arXiv preprint arXiv:1502.02072}, 2015.

\bibitem[Reed et~al.(2016)Reed, Akata, Yan, Logeswaran, Schiele, and
  Lee]{reed2016generative}
Reed, S., Akata, Z., Yan, X., Logeswaran, L., Schiele, B., and Lee, H.
\newblock Generative adversarial text to image synthesis.
\newblock \emph{arXiv preprint arXiv:1605.05396}, 2016.

\bibitem[Sandler et~al.(2018)Sandler, Howard, Zhu, Zhmoginov, and
  Chen]{sandler2018mobilenetv2}
Sandler, M., Howard, A., Zhu, M., Zhmoginov, A., and Chen, L.-C.
\newblock Mobilenetv2: Inverted residuals and linear bottlenecks.
\newblock In \emph{Proceedings of the IEEE Conference on Computer Vision and
  Pattern Recognition}, pp.\  4510--4520, 2018.

\bibitem[Such et~al.(2017)Such, Madhavan, Conti, Lehman, Stanley, and
  Clune]{such2017deep}
Such, F.~P., Madhavan, V., Conti, E., Lehman, J., Stanley, K.~O., and Clune, J.
\newblock Deep neuroevolution: Genetic algorithms are a competitive alternative
  for training deep neural networks for reinforcement learning.
\newblock \emph{arXiv preprint arXiv:1712.06567}, 2017.

\bibitem[Tan \& Le(2019)Tan and Le]{tan2019efficientnet}
Tan, M. and Le, Q.~V.
\newblock Efficientnet: Rethinking model scaling for convolutional neural
  networks.
\newblock \emph{arXiv preprint arXiv:1905.11946}, 2019.

\bibitem[Tan et~al.(2019)Tan, Chen, Pang, Vasudevan, Sandler, Howard, and
  Le]{mnasnet}
Tan, M., Chen, B., Pang, R., Vasudevan, V., Sandler, M., Howard, A., and Le,
  Q.~V.
\newblock Mnasnet: Platform-aware neural architecture search for mobile.
\newblock In \emph{Proceedings of the IEEE Conference on Computer Vision and
  Pattern Recognition}, pp.\  2820--2828, 2019.

\bibitem[Van~den Oord et~al.(2016)Van~den Oord, Kalchbrenner, Espeholt,
  Vinyals, Graves, et~al.]{van2016conditional}
Van~den Oord, A., Kalchbrenner, N., Espeholt, L., Vinyals, O., Graves, A.,
  et~al.
\newblock Conditional image generation with pixelcnn decoders.
\newblock In \emph{Advances in neural information processing systems}, pp.\
  4790--4798, 2016.

\bibitem[Wang et~al.(2019)Wang, Liu, Lin, Lin, and Han]{wang2019haq}
Wang, K., Liu, Z., Lin, Y., Lin, J., and Han, S.
\newblock Haq: Hardware-aware automated quantization with mixed precision.
\newblock In \emph{Proceedings of the IEEE Conference on Computer Vision and
  Pattern Recognition}, pp.\  8612--8620, 2019.

\bibitem[Wang et~al.(2018)Wang, Yu, Wu, Gu, Liu, Dong, Qiao, and
  Change~Loy]{wang2018esrgan}
Wang, X., Yu, K., Wu, S., Gu, J., Liu, Y., Dong, C., Qiao, Y., and Change~Loy,
  C.
\newblock Esrgan: Enhanced super-resolution generative adversarial networks.
\newblock In \emph{Proceedings of the European Conference on Computer Vision
  (ECCV)}, pp.\  0--0, 2018.

\bibitem[Xie et~al.(2017)Xie, Girshick, Doll{\'a}r, Tu, and
  He]{xie2017aggregated}
Xie, S., Girshick, R., Doll{\'a}r, P., Tu, Z., and He, K.
\newblock Aggregated residual transformations for deep neural networks.
\newblock In \emph{Proceedings of the IEEE conference on computer vision and
  pattern recognition}, pp.\  1492--1500, 2017.

\bibitem[Yoo et~al.(2016)Yoo, Kim, Park, Paek, and Kweon]{yoo2016pixel}
Yoo, D., Kim, N., Park, S., Paek, A.~S., and Kweon, I.~S.
\newblock Pixel-level domain transfer.
\newblock In \emph{European Conference on Computer Vision}, pp.\  517--532.
  Springer, 2016.

\bibitem[Yu et~al.(2017)Yu, Zhang, Wang, and Yu]{yu2017seqgan}
Yu, L., Zhang, W., Wang, J., and Yu, Y.
\newblock Seqgan: Sequence generative adversarial nets with policy gradient.
\newblock In \emph{Thirty-First AAAI Conference on Artificial Intelligence},
  2017.

\bibitem[Zhou et~al.(2017)Zhou, Yao, Guo, Xu, and Chen]{zhou2017incremental}
Zhou, A., Yao, A., Guo, Y., Xu, L., and Chen, Y.
\newblock Incremental network quantization: Towards lossless cnns with
  low-precision weights.
\newblock \emph{arXiv preprint arXiv:1702.03044}, 2017.

\bibitem[Zhou et~al.(2016)Zhou, Wu, Ni, Zhou, Wen, and Zou]{zhou2016dorefa}
Zhou, S., Wu, Y., Ni, Z., Zhou, X., Wen, H., and Zou, Y.
\newblock Dorefa-net: Training low bitwidth convolutional neural networks with
  low bitwidth gradients.
\newblock \emph{arXiv preprint arXiv:1606.06160}, 2016.

\end{thebibliography}
\bibliographystyle{sysml2019}



\end{document}